\documentclass[conference]{IEEEtran}
\IEEEoverridecommandlockouts
\usepackage{cite}
\usepackage{amsmath,amssymb,amsfonts}
\usepackage{algorithmic}
\usepackage{graphicx}
\usepackage{textcomp}
\usepackage{algorithm}
\def\BibTeX{{\rm B\kern-.05em{\sc i\kern-.025em b}\kern-.08em
    T\kern-.1667em\lower.7ex\hbox{E}\kern-.125emX}}

\newcommand{\argmin}{\mathop{\rm arg~min}\limits}

\begin{document}

%-----------------------------------------------------------------------------------------------
% title
%-----------------------------------------------------------------------------------------------
\title{
A Method Based on Convex Cone Model for Image-Set Classification with CNN Features
}
\author{
\IEEEauthorblockN{Naoya Sogi, Taku Nakayama and Kazuhiro Fukui}
\IEEEauthorblockA{
Graduate School of Systems and Information Engineering, University of Tsukuba, \\
1-1-1 Tennodai, Tsukuba, Ibaraki, 305-8573, Japan\\
Email: \{sogi, nakayama\}@cvlab.cs.tsukuba.ac.jp, kfukui@cs.tsukuba.ac.jp}
}

\maketitle
%-----------------------------------------------------------------------------------------------
% abst
%-----------------------------------------------------------------------------------------------
\begin{abstract}
In this paper, we propose a method for image-set classification based on convex cone models, focusing on the effectiveness of convolutional neural network (CNN) features as inputs. 
CNN features have non-negative values when using the rectified linear unit as an activation function.
This naturally leads us to model a set of CNN features by a convex cone and measure the geometric similarity of convex cones for classification. 
To establish this framework, we sequentially define multiple angles between two convex cones by repeating the alternating least squares method and then define the geometric similarity between the cones using the obtained angles.
Moreover, to enhance our method, we introduce a discriminant space, maximizing the between-class variance (gaps) and minimizes the within-class variance of the projected convex cones onto the discriminant space, similar to a Fisher discriminant analysis. 
Finally, classification is based on the similarity between projected convex cones. 
The effectiveness of the proposed method was demonstrated experimentally using a private, multi-view hand shape dataset and two public databases.
\end{abstract}


%-----------------------------------------------------------------------------------------------
% Sec1: Introduction
%-----------------------------------------------------------------------------------------------
\section{Introduction}

In this paper, we propose a method for image-set classification based on convex cone models that can deal with various types of features with non-negative constraint. We discuss the effectiveness of the combination with convolutional neural network (CNN) features extracted from a high-level hidden layer of a learned CNN.

For the last decade, image set-based classification methods \cite{msm,cmsm,kmsm,komsm,kcmsm,gds,imageset1}, and particularly subspace-based methods, such as the mutual subspace method (MSM)~\cite{msm} and constrained MSM (CMSM)~\cite{cmsm, gds}, have gaining substantial attention for various applications to multi-view images and videos, e.g., 3D object recognition and motion analysis, as they can handle a set of images effectively.  
In these methods, a set of images is compactly represented by a subspace in high dimensional vector space, where the subspace is generated by applying PCA to the image set without data centering. The classification of an input subspace is based on the canonical angles~\cite{canangles1,canangles2} between the input and each reference subspace, as the similarity index.

Conventional subspace-based methods assume a raw intensity vector or a hand-crafted feature as the input. Regarding more discriminant features, many recent studies have revealed that CNN features are effective inputs for various types of classifiers~\cite{cnnfeat1,cnnfeat_face,cnnfeat_sal,cnnfeat_maki}. 
Inspired by these results, subspace-based methods with CNN features have been proposed and have achieved high classification performance~\cite{cnn-msm}.

\begin{figure}[!t]
\centering
\includegraphics[width=8.5cm]{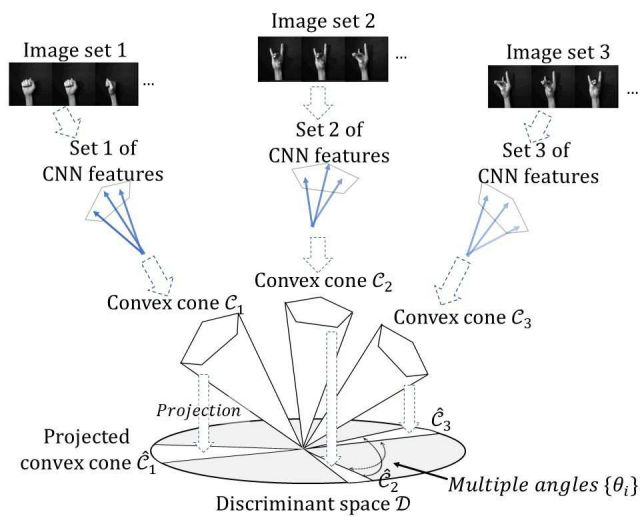}
\caption{Conceptual diagram of the proposed constrained mutual convex cone method (CMCM). First, a set of CNN features is extracted from an image set. Then, each set of CNN features is represented by a convex cone. After the convex cones are projected onto the discriminant space $\cal D$, the classification is performed by measuring similarity based on the angles $\{\theta_i\}$ between the two projected convex cones ${\hat{ \cal C}}_i$ and ${\hat{ \cal C}}_j$.}
\label{fig:abst}
\end{figure}

CNN feature vectors have only non-negative values when the rectified linear unit (ReLU)~\cite{relu} is used as an activation function. This characteristic does not allow the combination of CNN features with negative coefficients; accordingly, a set of CNN features forms a convex cone instead of a subspace in a high dimensional vector space, as described in Sec.\ref{sec:cone}.
For example, it is well known that a set of front-facing images under various illumination conditions forms a convex cone, referred to as an illumination cone~\cite{illumcone1,illumcone2}. Several previous studies have demonstrated the advantages of convex cone representation compared with subspace representation~\cite{cone,kcone}.
These advantages naturally motivated us to replace a subspace with a convex cone in models of a set of CNN features. 

In this framework, it is first necessary to consider how to calculate the geometric similarity between two convex cones. To this end, we define multiple angles between two convex cones by reference to the definition of the canonical angles~\cite{canangles1,canangles2} between two subspaces. 
Although the canonical angles between two subspaces can be analytically obtained from the orthonormal basis vectors of the two subspaces, the definition of angles between two convex cones is not trivial, as we need to consider the non-negative constraint.
In this paper, we define multiple angles between convex cones sequentially from the smallest to the largest by repeatedly applying the alternating least squares method~\cite{cone-cca}.
Then, the geometric similarity between two convex cones is defined based on the obtained angles.
We call the classification method using this similarity index the {\it mutual convex cone method} (MCM), corresponding to the mutual subspace method (MSM).

Moreover, to enhance the performance of the MCM, we introduce a discriminant space $\cal D$, which maximizes the between-class variance (gap) among convex cones projected onto the discriminant space and minimizes the within-class variance of the projected convex cones, similar to the Fisher discriminant analysis~\cite{fda}. 
The class separability can be increased by projecting the class of convex cones $\{{\cal C}_c\}$ onto the discriminant space $\cal D$, as shown in Fig.\ref{fig:abst}. As a result, the classification ability of MCM is enhanced, similar to that of the projection of class subspaces onto a generalized difference subspace (GDS) in CMSM \cite{gds}. Finally, we perform the classification using the angles between the projected convex cones $\{{\hat{\cal C}}_c\}$.
We call this enhanced method the ``{\it constrained mutual convex cone method} (CMCM)," corresponding to the constrained MSM (CMSM).

The main contributions of this paper are summarized as follows.
\begin{enumerate}
\item We introduce a convex cone representation to accurately and compactly represent a set of CNN features.
\item We introduce two novel mechanisms in our image set-based classification: a) multiple angles between two convex cones to measure similarity and b) a discriminant space to increase the class separability among convex cones.
\item We propose two novel image set-based classification methods, called the MCM and CMCM, based on convex cone representation and the discriminant space.
\end{enumerate}

The paper is organized as follows. 
In Section 2, we describe the algorithms in conventional methods, such as MSM and CMSM.
In Section 3, we describe the details of the proposed method.
In Section 4, we demonstrate the validity of the proposed method by classification experiments using a private database of multi-view hand shapes and two public datasets, i.e., ETH-80\cite{eth} and CMU face\cite{pie} datasets. 
Section 5 concludes the paper.

%-----------------------------------------------------------------------------------------------
% Sec2: Related work
%-----------------------------------------------------------------------------------------------
\section{Related work}
In this section, we first describe the algorithms for the MSM and CMSM, which are standard methods for image set classification. Then, we provide an overview of the concept of convex cones.

%-----------------------------------------
% subsec: MSM
\subsection{Mutual subspace method based on canonical angles}
%------------
% figure: Conceptual diagram of MSM
\begin{figure}[!t]
\centering
\includegraphics[width=7.5cm]{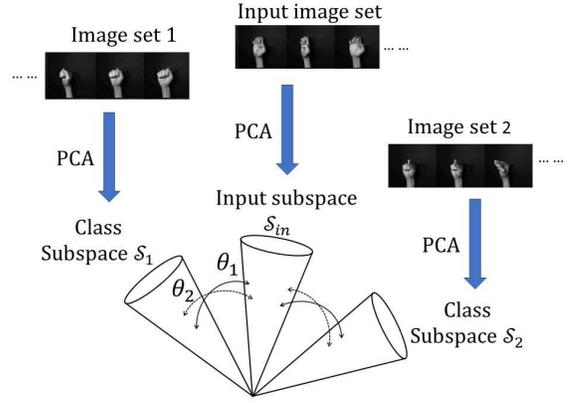}
\caption{Conceptual diagram of conventional MSM. Each image set is represented by a subspace, which is generated by applying the PCA to the set. In the classification, the similarity between two subspaces is measured based on the canonical angles between them. An input subspace is assigned to the class corresponding to the subspace with the greatest similarity.}
\label{fig:msm}
\end{figure}

MSM is a classifier based on canonical angles between two subspaces, where each subspace represents an image set.

\begin{figure}[bt]
  \begin{center}
    \includegraphics[trim=0 0 0 0,clip,width=55mm]{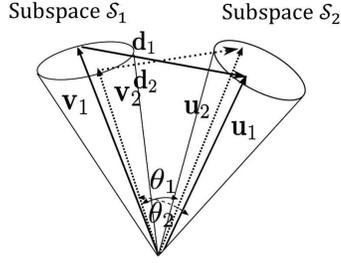}\vspace{2mm}\\
    \caption{Conceptual diagram of the canonical angles and canonical vectors. 1-st canonical vectors ${\bf u}_1,{\bf v}_1$ form the smallest angle $\theta_1$ between the subspaces. 2-nd canonical vectors ${\bf u}_2,{\bf v}_2$ form the smallest angle $\theta_2$ in a direction orthogonal to $\theta_1$.}
    \label{dspace}
  \end{center}
\end{figure}

Given $N_1$-dimensional subspace ${\cal S}_1$ and $N_2$-dimensional subspace ${\cal S}_2$ in $d$-dimensional vector space, where $N_1 \leq N_2$, the canonical angles $\{0\leq \theta_1,...,\theta_{N_1}\leq\frac{\pi}{2}\}$ between the ${\cal S}_1$ and ${\cal S}_2$ are recursively defined as follows\cite{canangles1,canangles2}:
\begin{align}
\cos{\theta_i}=\max_{{\bf u}\in {\cal S}_1}\max_{{\bf v}\in {\cal S}_2}{\bf u}^\mathrm{T}{\bf v}={\bf u}_i^\mathrm{T}{\bf v}_i, \\
s.t.\|{\bf u}_i\|_2=\|{\bf v}_i\|_2=1,{\bf u}_i^\mathrm{T}{\bf u}_j={\bf v}_i^\mathrm{T}{\bf v}_j=0,i\neq j, \nonumber
\end{align}
where ${\bf u}_i$ and ${\bf v}_i$ are the canonical vectors forming the $i$-th smallest canonical angle $\theta_i$ between ${\cal S}_1$ and ${\cal S}_2$. The $j$-th canonical angle $\theta_j$ is the smallest angle in the direction orthogonal to the canonical angles $\{\theta_k\}_{k=1}^{j-1}$ as shown in Fig.\ref{dspace}.

The canonical angles can be calculated from the orthogonal projection matrices onto subspaces ${\cal S}_1,{\cal S}_2$.
Let $\{{\bf \Phi}_i\}_{i=1}^{N_1}$ be basis vectors of ${\cal S}_1$ and $\{{\bf \Psi}_i\}_{i=1}^{N_2}$ be basis vectors of ${\cal S}_2$. 
The projection matrices ${\bf{P}}_1$ and ${\bf{P}}_2$ are calculated as $\sum_{i=1}^{N_1} {{\bf{\Phi}}_i} {{\bf{\Phi}}_i}^\mathrm{T}$ and $\sum_{i=1}^{N_2} {{\bf{\Psi}}_i}{{\bf{\Psi}}_i}^\mathrm{T}$, respectively.  
$\cos^2 \theta_i$ is the $i$-th largest eigenvalue of ${\bf P}_1^\mathrm{T}{\bf P}_2$ or ${\bf P}_2^\mathrm{T}{\bf P}_1$.
Alternatively, the canonical angles can be easily obtained by applying the SVD to the orthonormal basis vectors of the subspaces.

The geometric similarity between two subspaces ${\cal S}_1$ and ${\cal S}_2$ is defined by using the canonical angles as follows:
\begin{align}
sim({\cal S}_1, {\cal S}_2)=\frac{1}{N_1}\sum_{i=1}^{N_1}{\cos^2 \theta_i}.
\end{align}
In MSM, an input subspace ${\cal S}_{in}$ is classified by comparison with class subspaces $\{{\cal S}_c\}_{c=1}^C$ using this similarity.

%------------
% figure: Conceptual diagram of DS

%-----------------------------------------
% subsec: CMSM
\subsection{Constrained MSM}

\begin{figure}[!t]
\centering
\includegraphics[width=7cm]{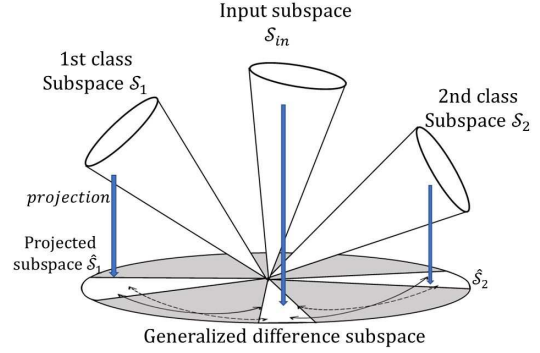}
\caption{Conceptual diagram of the constrained MSM (CMSM). By projecting class subspaces onto the generalized difference subspace, the separability between the classes is increased. By measuring the similarities among the projected subspaces using the canonical angles, the input subspace is assigned to either class 1 or 2.}
\label{fig:cmsm}
\end{figure}

The essence of the constrained MSM (CMSM) is the application of the MSM to a generalized difference subspace (GDS)~\cite{gds}, as shown in Fig.\ref{fig:cmsm}. GDS is designed to contain only difference components among subspaces $\{{\cal S}_c\}_{c=1}^C$. Thus, the projection of class subspaces onto GDS can increase the class separability among the class subspaces to substantially improve the classification ability of MSM \cite{gds}.

\subsection{Convex cone model}
\label{sec:cone}
In this subsection, we explain the definition of a convex cone and the projection of a vector onto a convex cone.
A convex cone $\cal{C}$ is defined by finite basis vectors $\{{\bf b}_i\}_{i=1}^r$ as follows:
\begin{equation}
\{{\bf a}\in{\cal C} | {\bf a} = \sum_{i=1}^r{w_i{\bf b}_i}, w_i\geq 0\}.
\end{equation}
As indicated by this definition, the difference between the concepts of a subspace and a convex cone is whether there are non-negative constraints on the combination coefficients $w_i$ or not.

Given a set of feature vectors $\{{\bf f}_i\}_{i=1}^N\in\mathbb{R}^d$, the basis vectors $\{{\bf b}_i\}_{i=1}^r$ of a convex cone representing a distribution of $\{{\bf f}_i\}$ can be obtained by non-negative matrix factorization (NMF)~\cite{nmf,nmfnnls}.
Let ${\bf F} = [{\bf f}_1 {\bf f}_2 ...{\bf f}_N]\in\mathbb{R}^{d\times N}$ and ${\bf B} = [{\bf b}_1 {\bf b}_2 ... {\bf b}_r]\in\mathbb{R}^{d\times r}$.
NMF generates the basis vectors ${\bf B}$ by solving the following optimization problem:
\begin{align}
\argmin_{{\bf B},{\bf W}} \|{\bf F-BW}\|_F~~ s.t.~~({\bf B})_{i,j},({\bf W})_{i,j}\geq 0,
\end{align}
where $\|\cdot\|_F$ denotes the Frobenius norm.
We use the alternating non-negativity-constrained least squares-based method~\cite{nmfnnls} to solve this problem. 

Although the basis vectors can be easily obtained by the NMF, the projection of a vector onto the convex cone is slightly complicated by the non-negative constraint on the coefficients.
In \cite{cone}, a vector ${\bf x}$ is projected onto the convex cone by applying the non-negative least squares method\cite{nnlsq} as follows:
%------
% Eq: nnls
\begin{align}
\argmin_{\{w_i\}}{\|{\bf x} - \sum_{i=1}^r{w_i{\bf b}_i}\|_2}~~ s.t.~~ w_i\geq 0.
\label{eq:nnls}
\end{align}
The projected vector ${\hat {\bf x}}$ is obtained as ${\hat {\bf x}} = \sum_{i=1}^r{w_i{\bf b}_i}$.

In the end, the angle $\theta$ between the convex cone and a vector ${\bf x}$ can be calculated as follows:
\begin{align}
\cos\theta =\frac{{\bf x}^{\mathrm{T}}{\hat {\bf x}}}{\|{\bf x}\|_2\|{\hat {\bf x}}\|_2}.
\end{align}

%------------
% figure: Conceptual diagram of CMSM

%-----------------------------------------------------------------------------------------------
% Sec3: Proposed method
%-----------------------------------------------------------------------------------------------
\section{Proposed method}
In this section, we explain the algorithms in the MCM and CMCM, after establishing the definition of geometric similarity between two convex cones. 

%-----------------------------------------
% subsec: Angles
\subsection{Geometric similarity between two convex cones}
\label{sec:angles}

We define the geometric similarity between two convex cones. To this end, we consider how to define multiple angles between two convex cones.
Two convex cones ${\cal C}_1$ and ${\cal C}_2$ are formed by basis vectors $\{{\bf b}^1_i\}_{i=1}^{N_1}\in \mathbb{R}^d$ and $\{{\bf b}^2_i\}_{i=1}^{N_2}\in \mathbb{R}^d$, respectively. Assume that $N_1 \leq N_2$ for convenience.
The angles between two convex cones cannot be obtained analytically like the canonical angles between two subspaces, as it is necessary to consider non-negative constraint. 
Alternatively, we find two vectors, ${\bf p} \in {\cal C}_1$ and ${\bf q} \in {\cal C}_2$, which are closest to each other. Then, we define the angle between the two convex cones as the angle formed by the two vectors. In this way, we sequentially define multiple angles from the smallest to the largeset, in order.

\begin{figure}[!t]
\centering
\includegraphics[width=7cm]{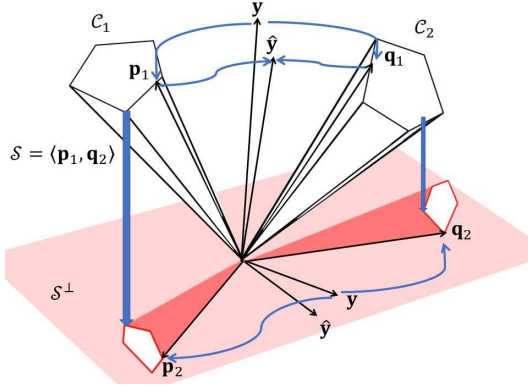}
\caption{Conceptual diagram of the procedure for searching pairs of vector $\{{\bf p}_i,{\bf q}_i\}$. The first pair of ${\bf p}_1$ and ${\bf q}_1$ can be found by the alternating least squares method. The second pair of ${\bf p}_2$ and ${\bf q}_2$ is obtained by searching the orthogonal complement space ${\cal S}^{\perp}$ of ${\cal S}=<{\bf p}_1,{\bf q}_1>.$}
\label{fig:near_vec}
\end{figure}

First, we search a pair of $d$-dimensional vectors ${\bf p}_1\in {\cal C}_1$ and ${\bf q}_1\in {\cal C}_2$, which have the maximum correlation, using the alternating least squares method (ALS)\cite{cone-cca}. The first angle $\theta_1$ is defined as the angle formed by ${\bf p}_1$ and ${\bf q}_1$. The pair of ${\bf p}_1$ and ${\bf q}_1$ can be found by using the following algorithm:

\vspace{2mm}
\noindent{\bf Algorithm to search for the pair ${\bf p}_1$ and ${\bf q}_1$}\\
Let ${\cal P}_1{\bf y}$ and ${\cal P}_2{\bf y}$ be the projections of a vector $\bf{y}$ onto ${\cal C}_1$ and ${\cal C}_2$, respectively. For the details of the projection, see Section~\ref{sec:cone}.
\begin{enumerate}
\item Randomly initialize $\bf{y}\in \mathbb{R}^d$.
\item ${\bf{p}}_1={\cal P}_1{\bf{y}} / \|{\cal P}_1{\bf{y}}\|_2$. 
\item ${\bf{q}}_1={\cal P}_2{\bf{y}} / \|{\cal P}_2{\bf{y}}\|_2$.
\item ${\hat {\bf y}} = ({\bf p}_1+{\bf q}_1) / 2$.
\item If $\|{\hat{\bf y}}-{\bf y}\|_2$ is sufficiently small, the procedure is completed. Otherwise, return to 2) setting ${\bf y}={\hat{\bf y}}$.
\item $\cos^2\theta_1={(\frac{{\bf p}_1^\mathrm{T}{\bf q}_1}{\|{\bf p}_1\|_2\|{\bf q}_1\|_2})^2}$.
\end{enumerate}
For the second angle $\theta_2$, we search for a pair of vectors ${\bf p}_2$ and ${\bf q}_2$ with the maximum correlation, but with the minimum correlation with ${\bf p}_1$ and ${\bf q}_1$. Such a pair can be found by applying ALS to the projected convex cones ${\cal C}_1$ and ${\cal C}_2$ on the orthogonal complement space ${\cal S}^{\perp}$ of the subspace ${\cal S}$ spanned by the vectors ${\bf p}_1$ and ${\bf q}_1$ as shown in Fig.\ref{fig:near_vec}. $\theta_2$ is formed by ${\bf p}_2$ and ${\bf q}_2$. In this way, we can obtain all of the pairs of vector ${\bf p}_i, {\bf q}_i$ forming $i$-th angle $\theta_i$, $(i=1,\dots,N_1)$.

With the resulting angles $\{\theta_i\}_{i=1}^{N_1}$, we define the geometrical similarity $sim$ between two convex cones ${\cal C}_1$ and ${\cal C}_2$ as follows:
%--------
% Eq: sim cone
\begin{align}
sim({\cal C}_1,{\cal C}_2) = \frac{1}{N_1}\sum_{i=1}^{N_1}{\cos^2\theta_i}.
\label{eq:sim_cone}
\end{align}

%------------
% figure: Nearest vectors procedure

\subsection{Mutual convex cone method}
\begin{figure}[!t]
\centering
\includegraphics[width=8cm]{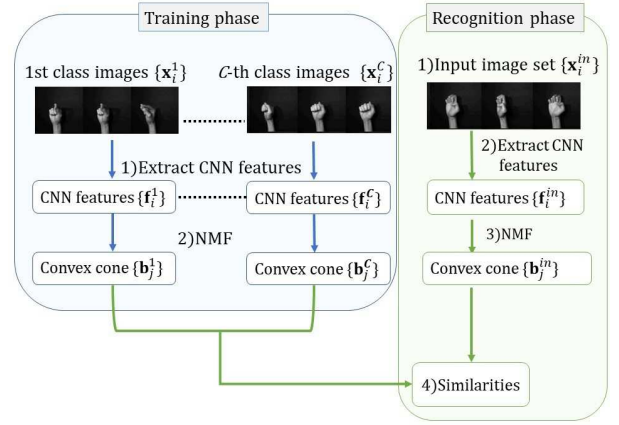}
\caption{Process flow of the proposed mutual convex cone method (MCM), which consists of a training phase and a testing phase.}
\label{fig:flow_mcm}
\end{figure}

The mutual convex cone method (MCM) classifies an input convex cone based on the similarities defined by Eq.(\ref{eq:sim_cone}) between the input and class convex cones.
MCM consists of two phases, a training phase and a recognition phase, as summarized in Fig.\ref{fig:flow_mcm}. 

Given $C$ class sets with $L$ images$ \{{\bf x}^c_i\}_{i=1}^L$.

\vspace{2mm}
\noindent{\bf Training Phase}
\begin{enumerate}
\item CNN features $\{{\bf f}_i^c\}$ are extracted from the images $\{{\bf x}_i^c\}$ of class $c$. Then, the extracted CNN features are normalized as $\{{\bf f}_i^c/\|{\bf f}_i^c\|_2\}$.
\item The basis vectors of $c$ class convex cone $\{{\bf b}_j^c\}$ are generated by applying NMF to the set of normalized CNN features.
\item $\{{\bf b}_{j}^c\}$ are registered as the reference convex cone of class $c$.
\item The above process is conducted for all $C$ classes.
\end{enumerate}

\vspace{2mm}
\noindent{\bf Recognition Phase}
\begin{enumerate}
\item A set of images $\{{\bf x}_i^{in}\}$ is input.
\item CNN features $\{{\bf f}_i^{in}\}$ are extracted from the images $\{{\bf x}_i^{in}\}$. Then, the extracted CNN features are normalized as $\{{\bf f}_i^{in}/\|{\bf f}_i^{in}\|_2\}$.
\item The basis vectors of the input convex cone $\{{\bf b}_j^{in}\}$ are generated by applying NMF to the input set of normalized CNN features.
\item The input image set $\{{\bf x}_i^{in}\}$ is classified based on the similarity (Eq.(\ref{eq:sim_cone})) between the input convex cone $\{{\bf b}_j^{in}\}$ and each $c$-th class reference convex cone $\{{\bf b}_j^c\}$.
\end{enumerate}

%------------
% figure: Framework of mcm

%-----------------------------------------
% subsec: GDSC
\subsection{Generation of discriminant space}
To enhance the class separability among multiple classes of convex cones, we introduce a discriminant space $\cal D$, which maximizes the between-class variance $\bf S_b$ and minimizes the within-class variance $\bf S_w$ for the convex cones projected on $\cal D$, similar to the Fisher discriminant analysis (FDA). In our method, the between class variance is replaced with gaps among convex cones.
We define these gaps as follows.
Let ${\cal C}_c$ be the $c$-th class convex cone with basis vectors $\{{\bf b}_i^c\}_{i=1}^{N_c}$, ${\cal P}_c$ be the projection operation of a vector onto ${\cal C}_c$, and $C$ be the number of the classes. 
We consider $C$ vectors $\{{\bf p}_1^c\}$, ($c=1,2,...,C)$ such that the sum of the correlation $\sum_{c{\neq}{c'}}{({\bf p}_1^c)^{\mathrm{T}}}{{\bf p}_1^{c'}/(\|{\bf p}_1^{c}\|_2\|{\bf p}_1^{c'}\|_2)}$ is maximum. Such a set of vectors can be obtained by using the following algorithm. This algorithm is almost the same as the generalized canonical correlation analysis (CCA) \cite{mcca1,mcca2}, except the non-negative least squares (LS) method is used instead of the standard LS method.

\vspace{2mm}
\noindent{\bf Procedure to search a set of first vectors $\{{\bf p}_1^c\}_{c=1}^C$}
\begin{enumerate}
\item Randomly initialize ${\bf y}_1$.
\item Project ${\bf y}_1$ onto each convex cone, and then normalize the projection as ${\bf p}_1^c={\cal P}_c{\bf y}_1/\|{\cal P}_c{\bf y}_1\|_2$.
\item ${\hat {\bf y}_1}=\sum_{c=1}^C{{\bf p}_1^c}/C$.
\item If $\|{\bf y}_1-{\hat{\bf y}}_1\|_2$ is sufficiently small, the procedure is completed. Otherwise, return to 2) setting ${\bf y}_1={\hat{\bf y}}_1$.
\end{enumerate}

Next, we search for a set of second vectors $\{{\bf p}_2^c\}$ with the maximum sum of the correlations under the constraint condition that they have the minimum correlation with the previously found $\{{\bf p}_1^c\}$.
To this end, we project the convex cones onto the orthogonal complement space of the vector ${\bf y}_1$. The second vectors $\{{\bf p}_2^c \}$ can be obtained by applying the above procedure to the projected convex cones.
In the following, a set of $j$-th vectors $\{{\bf p}_j^c\}$ can be sequentially obtained by applying the same procedure to the convex cones projected onto the orthogonal complement space of $\{{\bf y}_k\}_{k=1}^{j-1}$. In this way, we finally obtain the sets of $\{{\bf p}_j^c\}$.
With the sets of $\{{\bf p}_j^c\}$, we define a difference vector $\{{\bf d}_{j}^{c_1c_2}\}$ as follows:
\begin{align}
{\bf d}_{j}^{c_1c_2}= {\bf p}_j^{c_1}-{\bf p}_j^{c_2}.
\end{align}
Considering that each difference vector represents the gap between the two convex cones, we define $\bf S_b$ using these vectors as follows:
\begin{align}
{\bf S_b}=\sum_{j=1}^{\min(\{N_c\})}{\sum_{c_1=1}^{C-1}{\sum_{c_2=c_1+1}^{C}{{\bf d}_{j}^{c_1c_2}({\bf d}_{j}^{c_1c_2})^\mathrm{T}}}}.
\end{align}

Next, we define the within-class variance ${\bf S_w}$ using the basis vectors $\{{\bf b}_i^c\}$ for all of classes of convex cones as follows:
\begin{align}
{\bf S_w} = \sum_{c=1}^C{\sum_{i=1}^{N_c}{({\bf b}_i^c-{\bf \mu}_c)({\bf b}_i^c-{\bf \mu}_c)^\mathrm{T}}},
\end{align}
where ${\bf \mu}_c=\sum_{i=1}^{N_c}{{\bf b}_i^c}/{N_c}$.
Finally, the $N_d$-dimensional discriminant space $\cal D$ is spanned by $N_d$ eigenvectors $\{\phi_i\}_{i=1}^{N_d}$ corresponding to the $N_d$ largest eigenvalues $\{\gamma_i\}_{i=1}^{N_d}$ of the following eigenvalue problem:
\begin{align}
{\bf S_b}{\bf \phi}_i=\gamma_i{\bf S_w}{\bf \phi}_i.
\label{eigenpr}
\end{align}

\label{sec:gdsc}
%-----------------------------------------
% subsec: CCMSM
\subsection{Constrained mutual convex cone method}

%------------
% figure: Framework of cmcm
\begin{figure}[!t]
\centering
\includegraphics[width=8cm]{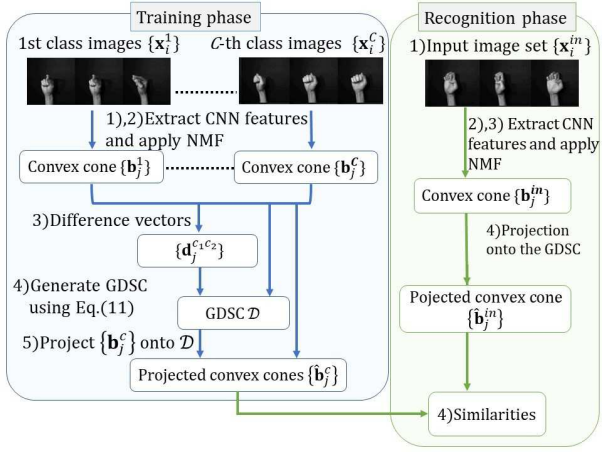}
\caption{Process flow of the proposed constrained MCM (CMCM). CMCM is an enhanced version of MSM with the projection of class subspaces onto the discriminant space $\cal D$.}
\label{fig:flow_cmcm}
\end{figure}

We construct the constrained MCM (CMCM) by incorporating the projection onto the discriminant space $\cal D$ into the MCM. 
CMCM consists of a training phase and a testing phase, as shown in Fig.\ref{fig:flow_cmcm}.
In the following, we explain each phase for the case in which $C$ classes have $L$ images$ \{{\bf x}^c_i\}_{i=1}^L$. 

\vspace{2mm}
\noindent{\bf Training Phase}
\begin{enumerate}
\item CNN features $\{{\bf f}_i^c\}$ are extracted from the images $\{{\bf x}_i^c\}$. Further, the extracted CNN features are normalized as $\{{\bf f}_i^c/\|{\bf f}_i^c\|_2\}$.
\item The basis vectors of the $c$ class convex cone $\{{\bf b}_j^c\}$ are generated by applying NMF to each class set of normalized CNN features.
\item Sets of difference vectors $\{{\bf d}_{j}^{c_1c_2}\}$ are generated according to the procedure described in section \ref{sec:gdsc}. 
\item The discriminant space ${\cal D}$ is generated by solving Eq.(\ref{eigenpr}) using $\{{\bf b}_j^c\}$ and $\{{\bf d}_{j}^{c_1c_2}\}$.
\item The basis vectors $\{{\bf b}_j^c\}$ are projected onto the discriminant space ${\cal D}$ and then the lengths of the projected basis vectors are normalized to 1.0.
A set of these  basis vectors $\{{\hat{\bf b}}_j^c\}$ forms the projected convex cone.
\item $\{{\hat{\bf b}}_{j}^c\}$ are registered as the reference convex cones of class $c$.
\end{enumerate}

\vspace{2mm}
\noindent{\bf Recognition Phase}
\begin{enumerate}
\item A set of images $\{{\bf x}_i^{in}\}$ is input.
\item CNN features $\{{\bf f}_i^{in}\}$ are extracted from the images $\{{\bf x}_i^{in}\}$. Further, the extracted CNN features are normalized as $\{{\bf f}_i^{in}/\|{\bf f}_i^{in}\|_2\}$ 
\item The basis vectors of a convex cone $\{{\bf b}_j^{in}\}$ are generated by applying NMF to the set of normalized CNN features.
\item The basis vectors $\{{\bf b}_j^{in}\}$ are projected onto the discriminant space ${\cal D}$ and then the lengths of the projected basis vectors are normalized to 1.0. The normalized projections are represented by $\{{\hat {\bf b}}_j^{in}\}$ 
\item The input set $\{{\bf x}_i^{in}\}$ is classified based on the similarity (Eq.(\ref{eq:sim_cone})) between the input convex cone $\{{\hat {\bf b}}_j^{in}\}$ and each class reference convex cone $\{{\hat {\bf b}}_j^c\}$.
\end{enumerate}

%-----------------------------------------------------------------------------------------------
% Sec4: Experiments
%-----------------------------------------------------------------------------------------------
\section{Evaluation Experiments}
In this section, we demonstrate the effectiveness of the proposed methods by comparative analyses of performance, including conventional subspace-based methods, MSM, and CMSM with CNN features. 
We used three databases: 1) multi-view hand shape dataset\cite{hand}, 2) ETH-80 dataset\cite{eth}, and 3) CMU Multi-PIE face dataset~\cite{pie}.
In the first three experiments, we conducted the classification using the three datasets with a sufficiently large number of training samples.
In the final experiment, we show the robustness of the proposed methods against small sample sizes (SSS), considering the situation in which few training samples can be used for learning. For the implementation of the methods, we used the NMF toolbox\cite{nmftool} and keras~\cite{keras}.
%-----------------------------------------
% subsec: Dataset

%------------
% figure: dataset
\begin{figure}[!t]
\centering
\includegraphics[width=6.5cm]{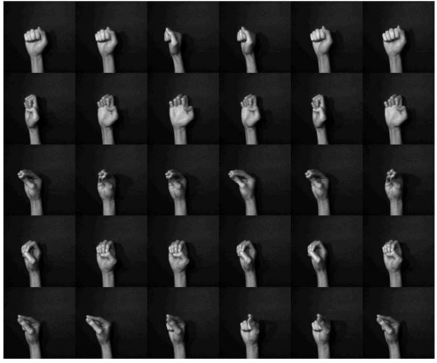}
\caption{Sample images of the multi-view hand shape dataset used in the experiments. Each row shows a hand shape from various viewpoints.}
\label{fig:dataset}
\end{figure}

%-----------------------------------------
% subsec: Recognition experiments
\subsection{Hand shape classification}

%-------------------
% subsubsec: details of the dataset
\subsubsection{Details of the dataset}
\label{sec:handdata}
The multi-view hand shape dataset consists of 30 classes of hand shapes.
Each class was determined from 100 subjects at a speed of 1 fps for 4 s using a multi-camera system equipped with seven synchronized cameras at intervals of 10 degrees.
During data collection, the subjects were asked to rotate their hands at a constant speed to increase the number of viewpoints. 
Figure \ref{fig:dataset} shows several sample images in the dataset.
The total number of images collected was 84000 (= 30 classes$\times$4 frames$\times$7 cameras $\times$100 subjects).

%-------------------
%subsubsec: experimental protocol
\subsubsection{Experimental protocol}
We used the same protocol as that described in ~\cite{gds}.
We randomly divided the subjects into two sets. 
One set was used for training, and the other was used for testing. That is, a reference convex cone for each hand shape was generated from a set of 1,400 (=7 cameras$\times$4 frames$\times$50 subjects) images.
As an input image set, we used 28 ($=$7 cameras$\times$4 frames) images.
The total number of convex cones used for testing was 1,500 (=30 shapes$\times$50 subjects).
We evaluated the classification performance of each method in terms of the average error rate (ER) of ten trials using randomly divided datasets.

We selected the parameters for the methods by cross validation using the training data.
For MSM and CMSM with CNN features, the dimensions of class, input subspaces, and GDS were set to 80, 5, and 200, respectively.
For conventional methods with raw images and FFT features, we used the same parameters as those in~\cite{gds}.
For MCM and CMCM, the numbers of basis vectors of class and input convex cones were set to 30 and 7, respectively. The dimension $N_d$ of the discriminant space $\cal D$ was set to 750.

To obtain CNN features under our experimental setting, we modified the original ResNet-50~\cite{resnet} trained by the Imagenet database\cite{imagenet} slightly for our experimental conditions. First, we replaced the final 1000-way fully connected (FC) layer of the original ResNet-50 with a 1024-way FC layer and applied the ReLU function. Further, we added a $\it class~number$-way FC layer with softmax behind the previous 1024-way FC layer.

Moreover, to extract more effective CNN features from our modified ResNet, we fine-tuned our ResNet using the learning set. A CNN feature vector was extracted from the 1024-way FC layer every time an image was input into our ResNet. As a result, the dimensionality $d$ of a CNN feature vector was 1024. 

In our fine-tuned CNN, an input image set was classified based on the average value of the output conviction degrees for each class from the last FC layer with softmax.
%------------
% Table: Error rate
\begin{table}[t]
\caption{Experimental results for the hand shape dataset.}
\begin{center}
\begin{tabular}{|c|c|c|}
\hline
Feature & Method & Error Rate(\%) \\ \hline \hline
		     & MSM\cite{gds}  & 22.55\\ \cline{2-3}
             & CMSM\cite{gds} & 17.12 \\ \cline{2-3}
Raw image    & KMSM\cite{gds}  & 12.52\\ \cline{2-3}
             & KCMSM\cite{gds} & 9.28 \\ \cline{2-3}
             & KOMSM\cite{gds} & 9.43 \\ \hline
             
		     & MSM\cite{gds}  & 17.69\\ \cline{2-3}
             & CMSM\cite{gds} & 8.78 \\ \cline{2-3}
FFT feature  & KMSM\cite{gds}  & 7.13\\ \cline{2-3}
             & KCMSM\cite{gds} & 5.84 \\ \cline{2-3}
             & KOMSM\cite{gds} & 5.66 \\ \hline
             
		     & softmax & 1.11 \\ \cline{2-3}
			 
CNN feature  & MSM  & 1.06\\ \cline{2-3}
	 		 & CMSM & 0.97\\ \hline  \hline
CNN feature  & MCM  & 1.47\\ \cline{2-3}
             & CMCM & \bf{0.92}\\\hline
\end{tabular}
\label{tab:exp1}
\end{center}
\end{table}
%-------------------
%subsubsec: Classification result
\subsubsection{Hand shape classification results}
Table \ref{tab:exp1} shows the average error rates for each method, including the proposed method. In the table, KCMSM indicates a non-linear extension of CMSM using the kernel trick\cite{kpca}.

We can see that the subspace- or convex cone-based methods with CNN features are significantly superior to methods with conventional features.
We can confirm the validity of CNN features. The results also indicate that a set of CNN features is more informative than the average value of the outputs from the last softmax layer.
When comparing the convex cone-based methods with the subspace-based methods, CMCM achieves the best performance.
This advantage suggests that a convex cone model is more suitable than a subspace model to represent a set of CNN features compactly and to effectively compare two sets.

%------------------------------------------
% subsec: ETH
\subsection{Object classification experiment}
We conducted an analysis of object classification using the ETH-80 dataset\cite{eth}.

\subsubsection{Details of the ETH-80 and experimental protocol}
The ETH-80 dataset consists of object images in eight different categories, captured from 41 viewpoints.
Each category has 10 kinds of object.

Five objects randomly sampled from each category set were used for training, and the remaining five objects were used for testing.
As an input image set, we used 41 images for each object.
We evaluated the classification performance of each method in terms of the average error rate (ER) of five trials using randomly divided datasets.

For MSM and CMSM, the dimensions of class subspaces, the input subspaces, and GDS were set to 55, 10, and 30, respectively.
For MCM and CMCM, the numbers of the basis vectors of class and input convex cones were set to 30 and 7, respectively. The dimension $N_d$ of the discriminant space $\cal D$ was set to 85.
We determined these dimensionalities by cross-validation using the training data.
CNN features were extracted from the fine-tuned ResNet under this experimental setting, according to the same procedure used in the previous experiments.

\subsubsection{Object classification result}
Table \ref{tab:exp_eth} shows the error rates for the different methods.
The CMCM exhibited the highest accuracy.
This result also supports the conclusion that a convex cone model is more appropriate to represent a set of CNN features than a subspace model.
In addition, we can confirm that the projection of the convex cones onto the discriminant space works well as a valid feature extraction. 

%------------
% Table: Error rate
\begin{table}[t]
\caption{Summary of results for the ETH-80 dataset.}
\begin{center}
\begin{tabular}{|c||c||c|c||c|c|}
\hline
Feature        &       \multicolumn{5}{|c|}{CNN feature} \\ \hline
Method         &   softmax  & MSM  & CMSM  & MCM  & CMCM \\ \hline
Error Rate(\%) &   {10.50}   & 12.00 & {11.00} & 9.50& {\bf 7.50 } \\\hline
\end{tabular}
\label{tab:exp_eth}
\end{center}
\end{table}

%-----------------------------------------
% subsec: face recognition
\subsection{Face classification experiment}
We conducted a face classification analysis using the CMU Multi-PIE dataset\cite{pie}.

\subsubsection{Details of the CMU dataset and experimental protocol}
The CMU Multi-PIE dataset consists of facial images of 337 different subjects captured from 15 viewpoints with 20 lighting conditions in 4 recording sessions.
In this experiment, we used images of 129 subjects captured from three viewpoints: front, left, and right. Thus, the total number of the images used for this experiment was 30960 ($=$129 subjects$\times$3 views$\times$20 illuminations$\times$4 sessions).

Two sessions were used for training, and the remaining two sessions were used for testing.
As an input image set, we used 10 randomly sampled images from an image set for each subject.
For MSM and CMSM, the dimensions of class, input subspaces, and GDS were set to 20, 5, and 520, respectively.
For MCM and CMCM, the numbers of the basis vectors of class and input convex cones were set to 20 and 5, respectively. The dimension $N_d$ of the discriminant space $\cal D$ was set to 530.
We determined these parameters by cross validation.
We used CNN features extracted from the fine-tuned ResNet using the training data, following the experimental setting.

\subsubsection{Face classification results}

Table \ref{tab:exp3} shows the error rates for the different methods.
The CMCM exhibited the highest performance, while the performance of the MCM was the lowest.
This result supports the validity of the projection onto the discriminant space as a feature extraction.
This implies that the gaps between convex cones capture useful geometrical information to enhance the class separability among all classes of convex cones.

%------------
% Table: Error rate
\begin{table}[t]
\caption{Summary of results for the CMU PIE dataset.}
\begin{center}
\begin{tabular}{|c||c||c|c||c|c|}
\hline
Feature        &    \multicolumn{5}{|c|}{CNN feature} \\ \hline
Method         &   softmax  & MSM  & CMSM  & MCM  & CMCM \\ \hline
Error Rate(\%) &   4.22    & 4.01 & 3.75  & 4.84 & {\bf 3.55 } \\\hline
\end{tabular}
\label{tab:exp3}
\end{center}
\end{table}

%------------
% figure: Result of ER in SSS problem.
\begin{figure}[!t]
\centering
\includegraphics[width=9cm]{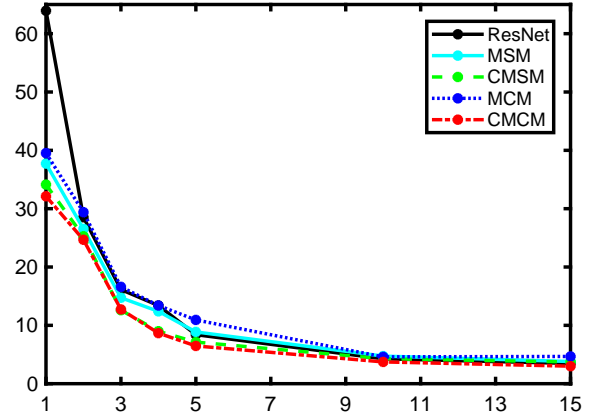}
\caption{Change in the error rates ($\%$) against the number of training subjects. The horizontal axis denotes the number of training subjects and the vertical axis denotes the error rates.}
\label{fig:exp2_er}
\end{figure}

%-----------------------------------------
% subsec: small sample recognition
\subsection{Robustness against limited training data}
A major issue with deep neural networks is the requirement for a large quantity of training samples to accurately learn the networks. Therefore, the robustness against a small sample size (SSS) is a necessary characteristic for effective methods using CNN features in practical applications.
In this experiment, we evaluated the robustness of the different methods against SSS.

%-------------------
% subsubsec: protocol of small sample recognition 
\subsubsection{Experimental protocol}
In this experiment, we used the hand shape dataset described in section \ref{sec:handdata}.
The dataset was divided into two sets in the same manner used for the previous experiment.
One set was used for training and another was used for testing.
We evaluated the performances of the methods by setting the numbers of subjects used for training to 1, 2, 3, 4, 5, 10, and 15.
In each case, the total number of training images was 30 classes$\times$7 cameras$\times$4 frames$\times$ $n$ subjects, ($n=1, 2, 3, 4, 5, 10, 15$).
As an input image set, we used 28 (=7 cameras $\times$4 frames) images, as in the previous experiment. Thus, the total number of convex cones for testing was 1500 (=30 classes$\times$50 subjects).

The parameters for the methods were determined by cross validation using training images.
For MSM and CMSM, the dimensions of class, input subspaces, and GDS were set to 25, 7, and 725, respectively.
For MCM and CMCM, the numbers of the basis vectors of class and input convex cones were set to 30 and 7, respectively. The dimension $N_d$ of the discriminant space $\cal D$ was set to 800.

To extract CNN features from the images, we used the fine-tuned ResNet by using the training images under the experimental conditions.

%-------------------
% Experiment result
\subsubsection{Summary of results}
Figure \ref{fig:exp2_er} shows the error rates in terms of the number $N$ of training subjects.
As shown in the figure, we can see that the overall performance of CMCM was better than that of the other methods. In particular, CMCM works well when the number of training subjects $N$ is small.
For example, when $N$ is 1, CMSM and CMCM achieve an error rate of about half that for softmax.
Moreover, CMCM improves the performances of the subspace-based methods, MSM and CMSM.
This further indicates that the convex cone method can represent the distribution of a set of CNN features more accurately than the subspace-based methods.

%-----------------------------------------------------------------------------------------------
% Sec5: conclusion
%-----------------------------------------------------------------------------------------------
\section{Conclusion}
In this paper, we proposed a method based on the convex cone model for image-set classification, referred to as the constrained mutual convex cone method (CMCM).
We discussed a combination of the proposed method and CNN features, though our method can be applied to various types of features with non-negative constraint.

The main contributions of this paper are 1) the introduction of a convex cone model to represent  a set of CNN features compactly and accurately, 2) the definition of the geometrical similarity of two convex cones based on the angles between them, which are obtained by the alternating least squares method, 3) the proposal of a method, i.e., MCM, for classifying convex cones using the angles as the similarity index, 4) the introduction of a discriminant space that maximizes between-class variance (gaps) between convex cones and minimizes within-class variance, 5) the proposal of the constrained MCM (CMCM), which incorporates the above projection into the MCM.

We demonstrated the validity of our methods by three experiments using the multi-view hand shape dataset, the CMU PIE dataset, and ETH-80.
In the future, we will evaluate the introduction of non-linear mapping by a kernel function into the proposed methods.

%-----------------------------------------------------------------------------------------------
% Sec4: Acknowledgement
%-----------------------------------------------------------------------------------------------
\section*{Acknowledgment}
This work was partially supported by JSPS KAKENHI Grant Number JP16H02842.

%-----------------------------------------------------------------------------------------------
% Reference
%-----------------------------------------------------------------------------------------------
\bibliographystyle{IEEEtran}
\bibliography{forijcnn}

\end{document}